\documentclass[a4paper, 10pt, conference]{ieeeconf}
\IEEEoverridecommandlockouts
\let\labelindent\relax
\usepackage{epsfig}
\usepackage{graphicx}
\usepackage{hyperref}
\usepackage{color}
\usepackage[symbol]{footmisc}

\newcommand{\realfield}[1]{\hbox{I \kern -.25em R}^{#1}}
\newcommand {\mb}[1]{\mathbf{#1}}
\newcommand {\bs}[1]{\boldsymbol{#1}}

\newcommand{\T}{^{\mathrm{T}}}

\usepackage[%
font=small,
labelfont=bf,
singlelinecheck=off]{caption}
\usepackage{array}
\newcolumntype{P}[1]{>{\centering\arraybackslash}p{#1}}
\newcolumntype{M}[1]{>{\centering\arraybackslash}m{#1}}
\usepackage{lipsum}
\usepackage{comment}
\usepackage{textcomp}
\usepackage{gensymb}
\usepackage{amsmath}
\usepackage{amssymb}
\usepackage{mathtools}
\usepackage{commath}
\usepackage{etoolbox}
\usepackage{caption}
\usepackage{subcaption}
\usepackage{siunitx}
\usepackage[compress]{cite}
\usepackage{cuted}
\usepackage{float}
\floatstyle{plaintop}
\restylefloat{table}
\usepackage{enumitem}
\usepackage{arydshln}
\graphicspath{{figures/}}

\usepackage[textsize=scriptsize, bordercolor=white,backgroundcolor=gray!30,linecolor=black,colorinlistoftodos]{todonotes} 

\usepackage[normalem]{ulem}
\usepackage{soul} 

\usepackage{xcolor}

\newcommand{\cusst}[1]{{}}
\usepackage{framed}
\usepackage{algorithm}
\usepackage{algpseudocode}
\usepackage{setspace}
\usepackage{multirow}
\graphicspath{{figures/}}
\begin{document}
\title{A Bimanual Teleoperation Framework for Light Duty Underwater Vehicle-Manipulator Systems}

\author{Justin~Sitler
\and Srikarran~Sowrirajan
\and Brendan~Englot
\and Long~Wang
        \thanks{J. Sitler, S. Sowrirajan, B. Englot, and L. Wang are with the Department of Mechanical Engineering, Stevens Institute of Technology, New Jersey, NJ, 07030, USA. Email:\texttt{\{jsitler1,ssowrira,benglot,lwang4\}@stevens.edu}}
        \thanks{This research was supported in part by USDA-NIFA Grant No. 2021-67022-35977 and NSF Grant CMMI-2138896.}
}

\maketitle
\colorlet{shadecolor}{yellow}
\setstcolor{red}
\begin{abstract}

In an effort to lower the barrier to entry in underwater manipulation, this paper presents an open-source, user-friendly framework for bimanual teleoperation of a light-duty underwater vehicle-manipulator system (UVMS). This framework allows for the control of the vehicle along with two manipulators and their end-effectors using two low-cost haptic devices. 

The UVMS kinematics are derived in order to create an independent resolved motion rate controller for each manipulator, which optimally controls the joint positions to achieve a desired end-effector pose. This desired pose is computed in real-time using a teleoperation controller developed to process the dual haptic device input from the user. A physics-based simulation environment is used to implement this framework for two example tasks as well as provide data for error analysis of user commands\footnote{\url{https://youtu.be/5pHZ5DaIIY4}}. The first task illustrates the functionality of the framework through motion control of the vehicle and manipulators using only the haptic devices. The second task is to grasp an object using both manipulators simultaneously, demonstrating precision and coordination using the framework. The framework code is available at \url{https://github.com/stevens-armlab/uvms_bimanual_sim}.

\end{abstract}

\section{Introduction}
\label{section:introduction}

\begin{figure}
    \centering
    \includegraphics[width=\linewidth]{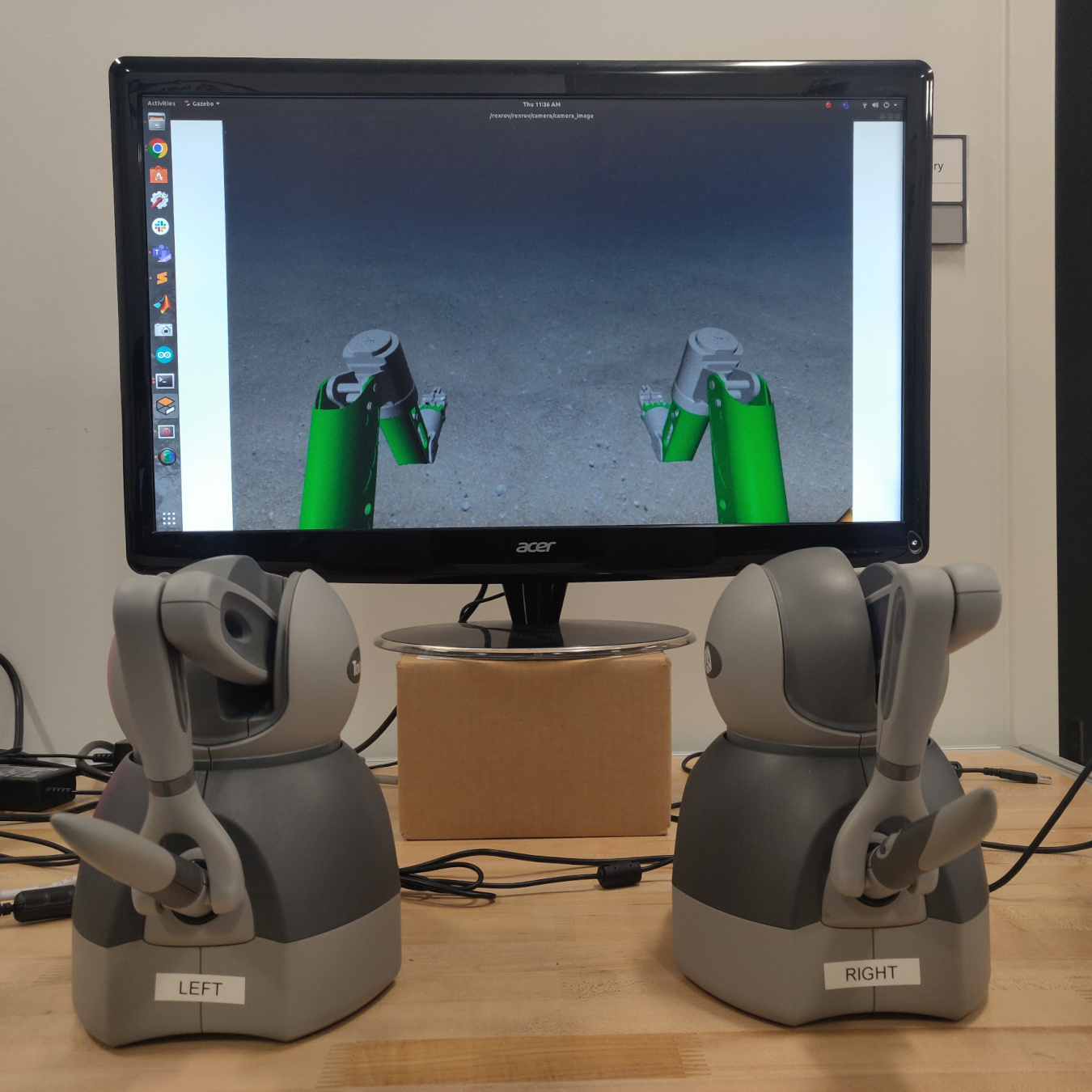}
    \caption{User workspace for bimanual teleoperation framework}
    \label{fig:placeholder}
\end{figure}

Underwater robotics has many useful applications such as exploration \cite{trotter2019}, inspection \cite{fernandes2015}, and aquaculture \cite{rundtop2016}. However, this research has typically involved large and expensive robots, such as the Girona 500 I-AUV \cite{ribas2011, ribas2012}, SAUVIM \cite{yuh1998, marani2009}, or REMUS \cite{stokey2005, jaffre2019}. The existing hardware is a huge barrier to entry for research in this field, and as such there has been some recent work developing light-duty platforms such as the Blue Robotics BlueROV2 \cite{bluerobotics}, which has been used for research in works by Marais et. al. \cite{marais2021, marais2022} as well as McConnell et. al. \cite{mcconnell2020, mcconnell2021, mcconnell2022}. 
Underwater robots are typically free-floating vehicles either teleoperated by a pilot \cite{moniruzzaman2022, khatib2016, jakuba2018} and referred to as remotely operated vehicles (ROVs), or autonomously controlled \cite{manley2018, valavanis1997} and referred to as autonomous underwater vehicles (AUVs). When equipped with a manipulator to perform intervention tasks, these robots are referred to as underwater vehicle-manipulator systems (UVMS). 

The control of a UVMS is complicated due to the need for separate vehicle and manipulator control. A work package for teleoperation via wireless acoustic subsea communication is presented in \cite{sayers1996}, which presents a graphical user interface (GUI) with virtual and camera feedback to visualize the deployed system. The user can control the UVMS with a master device or use a mouse to send commands via the GUI. The system employs synthetic fixtures to assist the user through force and visual cues \cite{sayers1994}. A user-friendly GUI was developed for semi-autonomous control of a BlueROV2 for floating debris collection in \cite{shirakura2021}, in which vehicle commands are generated via interaction with the GUI.

Teleoperation of manipulator systems enables an operator to communicate specific maneuvers to a follower robot for execution. Telerobotics is popular in applications such as robotic surgery, where skilled surgeons conduct minimally-invasive surgery through visual feedback as presented in \cite{burgner2014}. The 3D Systems Touch,  formerly Phantom Omni, is a haptic device that captures pose information from the operator for teleoperation \cite{3dstouch}. Bimanual teleoperation using the haptic device enables surgeons to perform base skull surgery, where haptic feedback offers an added dimension of realism to the surgeons, as presented in \cite{basdogan2004}. 

To lower the barrier to entry into underwater robotics we propose a simple, user-friendly framework for bimanual teleoperation of a light duty UVMS, an overview of which can be seen in Fig. \ref{fig:placeholder}. This paper derives the kinematic model for UVMS telemanipulation and implements a resolved motion rate controller. The mapping between the leader haptic device frames and the follower arm frame is also derived to generate desired manipulator poses from the operator. In addition, the framework is implemented in a physics simulation environment to validate its performance in two different tasks. The first task is manual piloting of the vehicle, manipulators, and end-effectors to a desired pre-grasp pose. The second task involves grasping a small object using both manipulators simultaneously. 

Our framework is unique from prior examples in that it uses two low-cost haptic devices for bimanual teleoperation instead of the expensive hardware and complex consoles used in the control of other bimanual UVMSs such as the Ocean One Avatar \cite{khatib2016} or the Aquanaut \cite{manley2018}. The framework is also developed using open source tools such as Gazebo \cite{koenig2004}, the UUV Simulator \cite{manhaes2016}, the Collaborative Robotics Toolkit (CRTK) \cite{CRTK2020}, and our own open source code \cite{uvms_bimanual_sim}, so any researcher can easily adapt the framework to their own needs. In addition, we demonstrate a novel vehicle control method using the haptic devices to send vehicle commands, allowing for a single user to quickly and easily switch between vehicle, manipulator, and end-effector control. 


The outline of this paper is as follows. Section \ref{section:uvms-kinematics} derives the kinematics for a generalized UVMS and the resolved motion rate controller which generates real-time manipulator joint commands from the desired end-effector pose. Section \ref{section:trajectory-mapping} introduces the mapping between leader-follower devices, from which end-effector commands are generated. Section \ref{section:simulation-model} discusses the simulation environment and UVMS models on which the bimanual control is implemented. Section \ref{section:simulation-results} presents two example tasks executed in simulation using the bimanual controller. Finally, Section \ref{section:conclusion} reflects on future improvements that could build on the work presented in this paper.

\section{UVMS Kinematics}
\label{section:uvms-kinematics}

\subsection{Forward Kinematics}

\begin{figure}
    \includegraphics[width=\linewidth]{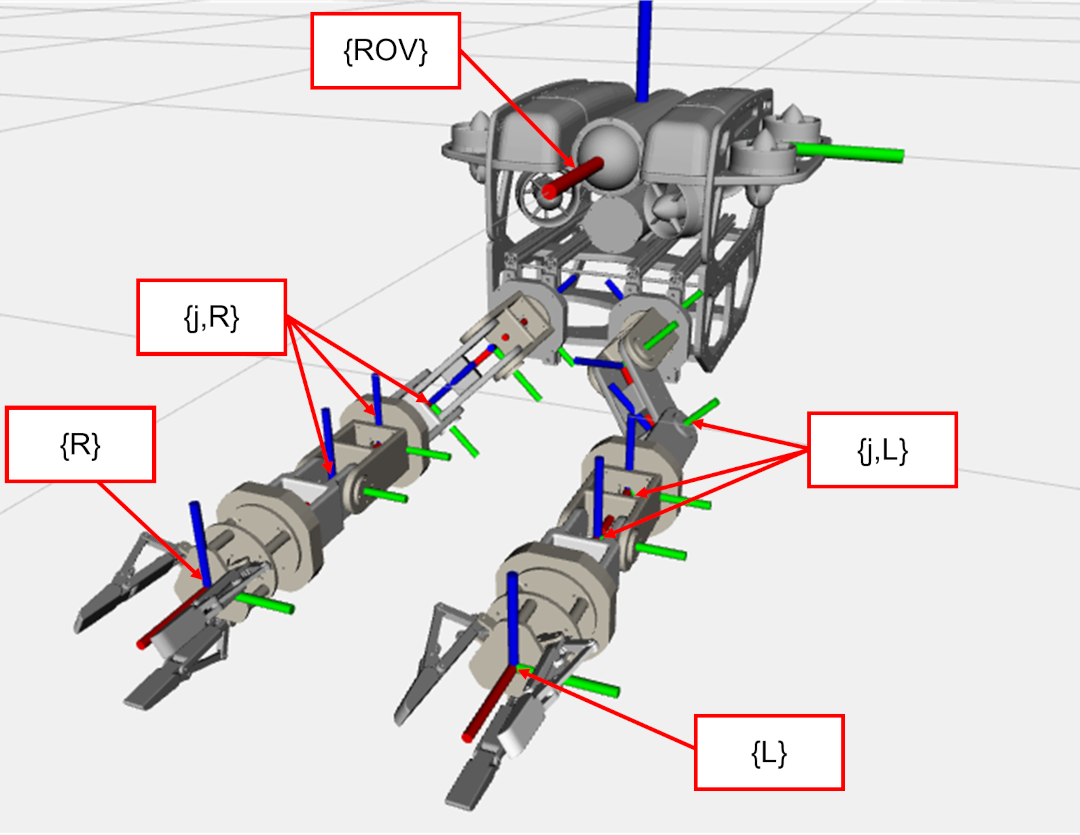}
    \caption{Coordinate frames used in the kinematic model. Frame indices $\{j,i\}$ begin with $j=1$ at the base of the manipulator and $j=6$ equivalent to the end-effector frames $\{L\}$ and $\{R\}$}
    \label{fig:UVMS-frames}
\end{figure}

A kinematic model of the UVMS is a prerequisite for teleoperational control. In this section, the direct kinematics and Jacobian matrix are derived in general for a serial manipulator of 6 degrees of freedom (DoF), denoted by the subscript $i=L,R$ representing the left and right manipulators respectively. Figure~\ref{fig:UVMS-frames} shows the coordinate frames of a forward-facing configuration for two serial manipulators on the BlueROV2 Heavy platform. 

For convenience, the kinematics of the UVMS are defined relative to the vehicle frame $\{ROV\}$ except where otherwise noted. The end-effector pose of manipulator $i$ is defined as:
\begin{equation}
    \mb{x}_i \triangleq \left(\mb{p}_i, \; \mb{R}_i \right)\;,
\end{equation}
\noindent where $\mb{p}_i$ is the position of the end-effector and $\mb{R}_i$ is the $3\times3$ rotation matrix representing the orientation of the end-effector. The end-effector pose can also be represented as the $4\times4$ transformation matrix:
\begin{equation}
    \left(\prescript{ROV}{}{\mb{T}_{ee}}\right)_i \triangleq \begin{bmatrix}
        \mb{R}_i & \mb{p}_i \\
        \mb{0}_{1\times3} & 1
    \end{bmatrix}.
\end{equation}

\noindent The joint variables of each manipulator are independent from each other and are defined as: 
\begin{equation}
    \mb{q}_i \triangleq \begin{bmatrix} q_{1,i} & q_{2,i} & \cdots & q_{6,i} \end{bmatrix}\T\;.
\end{equation}

\noindent The position and orientation of the end-effector can be found sequentially:
\begin{equation}
    (\prescript{ROV}{}{\mb{T}_{ee}})_i = \prescript{ROV}{}{\mb{T}_{1,i}} \prescript{1,i}{}{\mb{T}_{2,i}} \prescript{2,i}{}{\mb{T}_{3,i}} \prescript{3,i}{}{\mb{T}_{4,i}} \prescript{4,i}{}{\mb{T}_{5,i}} \prescript{5,i}{}{\mb{T}_{6,i}}\;,
\end{equation}
where $\prescript{j-1,i}{}{\mb{T}_{j,i}}$ represents the relative transformation between frames $\{j-1,i\}$ and $\{j,i\}$ as defined in Fig. \ref{fig:UVMS-frames}.  The Jacobian for manipulator $i$ relates the end-effector twist $\bs{\xi}_i$ to the joint velocities $\dot{\mb{q}}_i$ and is derived geometrically:
\begin{align}
    &\bs{\xi}_i \triangleq \mb{J}_i \dot{\mb{q}},  \quad\mb{J}_i= \begin{bmatrix} \mb{J}_{p,i}\\ \mb{J}_{o,i} \end{bmatrix}, \quad \bs{\xi}_i \triangleq [{\mb{v}_i}\T, \; {\bs{\omega}_i}\T]\T \label{eq:twist}\\
    &\mb{J}_{p,i} = \begin{bmatrix} \hat{\mb{s}}_{1,i} \times \left(\mb{p}_i - \mb{p}_{1,i}\right) & \cdots & \hat{\mb{s}}_{6,i} \times \left(\mb{p}_i - \mb{p}_{6,i}\right)\end{bmatrix}\\
    &\mb{J}_{o,i} = \begin{bmatrix} \hat{\mb{s}}_{1,i} & \cdots & \hat{\mb{s}}_{6,i} \end{bmatrix}\;,
\end{align}
where $\mb{p}_{j,i}$ represents the position of frame $\{j,i\}$ and $\hat{\mb{s}}_{j,i}\footnote{$\hat{(\cdot)}$ denotes a unit vector.}$ represents the axis of rotation of the corresponding joint.

\subsection{Resolved Motion Rate Control}
Control of both manipulators is achieved by a resolved motion rate algorithm that separately tracks joint commands. This was first proposed in \cite{whitney1969} and has also been adapted by Marani et. al. for singularity avoidance in control of the SAUVIM underwater robot's manipulator \cite{marani2002}. The desired pose, denoted as $\mb{x}_{d,i} \triangleq \left(\mb{p}_{d,i}, \; \mb{R}_{d,i}\right)$, is specified in real-time by operator input as described in Section \ref{section:trajectory-mapping}. Therefore, the position error $\mb{e}_{p,i}$ of manipulator $i$ between the current pose $\mb{x}_{i}$ is given by:
\begin{equation}
    \mb{e}_{\mb{p},i} = \mb{p}_{d, i} - \mb{p}_i.
\end{equation}

The orientation error $\mb{e}_{\bs{\mu},i}$ of manipulator $i$ is found by first deriving the relative error rotation matrix $\mb{R}_{e,i}$ between the current orientation $\mb{R}_i$ and desired orientation $\mb{R}_{d,i}$:
\begin{align}
    &\mb{R}_{e,i} = \mb{R}_{d, i}  \; {\mb{R}_{i}}\T \\
    &\theta_i = \arccos{\frac{\text{tr}(\mb{R}_{e,i}) - 1}{2}}\\
    &\mb{e}_{\bs{\mu},i} = \frac{\theta_i}{2 \sin{\theta_i}} \begin{bmatrix}
        \mb{R}_{{e,i}\;(3,2)} - \mb{R}_{{e,i}\;(2,3)}\\
        \mb{R}_{{e,i}\;(1,3)} - \mb{R}_{{e,i}\;(3,1)}\\
        \mb{R}_{{e,i}\;(2,1)} - \mb{R}_{{e,i}\;(1,2)}
    \end{bmatrix} .
\end{align}

The position and orientation errors are used in resolved motion rate control to calculate the desired end-effector twist $\bs{\xi}$. As the end-effector approaches the desired position and the errors approach zero, the magnitude of the end-effector twist decreases:
\begin{align}
    &v_\text{mag} = \begin{cases}
        v_\text{M} & \lVert \mb{e}_{\mb{p},i} \rVert > \frac{\lambda_\mb{p}}{\delta_{\mb{p}}} \\
        v_\text{m} + (v_\text{M} - v_\text{m})\frac{\lVert \mb{e}_{\mb{p},i} \rVert - \delta_\mb{p}}{\delta_\mb{p}(\lambda_\mb{p} - 1)} & \lVert \mb{e}_{\mb{p},i} \rVert \le \frac{\lambda_\mb{p}}{\delta_\mb{p}}
    \end{cases}\\
    &\omega_\text{mag} = \begin{cases}
        \omega_\text{M} & \lVert \mb{e}_{\bs{\mu},i} \rVert > \frac{\lambda_{\bs{\mu}}}{\delta_{\bs{\mu}}} \\
        \omega_\text{m} + (\omega_\text{M} - \omega_\text{m})\frac{\lVert \mb{e}_{\bs{\mu},i} \rVert - \delta_{\bs{\mu}}}{\delta_{\bs{\mu}}(\lambda_{\bs{\mu}} - 1)} & \lVert \mb{e}_{\bs{\mu},i} \rVert \le \frac{\lambda_{\bs{\mu}}}{\delta_{\bs{\mu}}}
    \end{cases}\\
    &\mb{v}_i = v_\text{mag} \frac{\mb{e}_{\mb{p},i}}{\lVert \mb{e}_{\mb{p},i} \rVert} \\
    &\bs{\omega}_i = \omega_\text{mag} \frac{\mb{e}_{\bs{\mu},i}}{\lVert \mb{e}_{\bs{\mu},i} \rVert} \;,
\end{align}

\noindent where $v_\text{M}$ and $v_\text{m}$ represent the maximum and minimum linear velocity of the end-effector; $\omega_\text{M}$ and $\omega_\text{m}$ represent the maximum and minimum angular velocity of the end-effector; $\delta_\mb{p}$ and $\delta_{\bs{\mu}}$ represent the allowable position and orientation error thresholds; and $\lambda_\mb{p}$ and $\lambda_{\bs{\mu}}$ represent the relative error radius in which the end-effector switches from a constant (resolved) rate to a reduced proportional rate. The joint velocities can be calculated from the twist defined in Eq. \ref{eq:twist}:
\begin{equation}
    \dot{\mb{q}}_i = \mb{J}_i^{+} \footnote{The pseudo-inverse of the Jacobian is used in the general case; however, in this paper the manipulator is presumed to have 6 DoFs, in which case the Jacobian is square and the standard inverse can be used instead.} \bs{\xi}_i \;,
\end{equation}
\noindent where $\mb{J}_i^{+}$ denotes Moore-Penrose pseudo-inverse of $\mb{J}_i$. The manipulator joint commands can then be found iteratively:
\begin{equation}
    \mb{q}_i = \mb{q}_i + \dot{\mb{q}}_i \text{d} t \;.
\end{equation}

\section{Teleoperation}
\label{section:trajectory-mapping}

Two 3D Systems Touch haptic devices are designated as the leaders for bimanual teleoperation of the UVMS. The devices each have 6 degrees of freedom of motion and are equipped with 2 buttons on the stylus. The software package presented in \cite{CRTK2020} provides the necessary APIs to communicate with the devices using Robot Operating System (ROS) \cite{Quigley09}, and retrieve information such as the pose of the stylus and the state of the buttons. The teleoperation framework can be separated into commanding manipulator trajectory and vehicle teleoperation. The type of teleoperation command executed by the UVMS is determined by the sequence or combination of buttons pressed on the stylus of one or both devices. The grippers on the left or right manipulators are toggled between open and closed states by clicking both buttons simultaneously on the corresponding left or right haptic device. 

\begin{figure*}
    \centering
    \includegraphics[width=\linewidth]{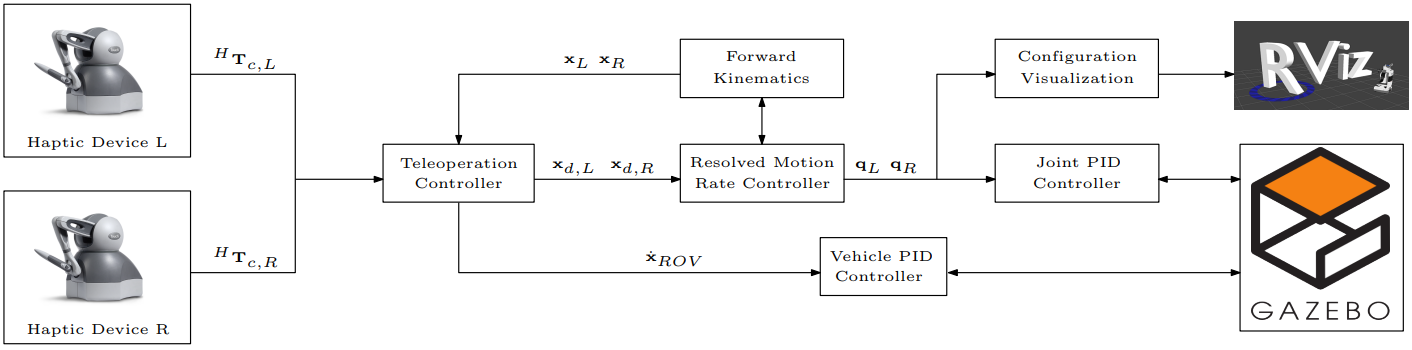}
    \caption{Control architecture for bimanual teleoperation of UVMS}
    \label{fig:rr-schematic}
\end{figure*}

\begin{figure}
    \captionsetup[subfigure]{justification=centering}
    \begin{subfigure}{\linewidth}
        \centering
        \includegraphics[width=0.7\linewidth]{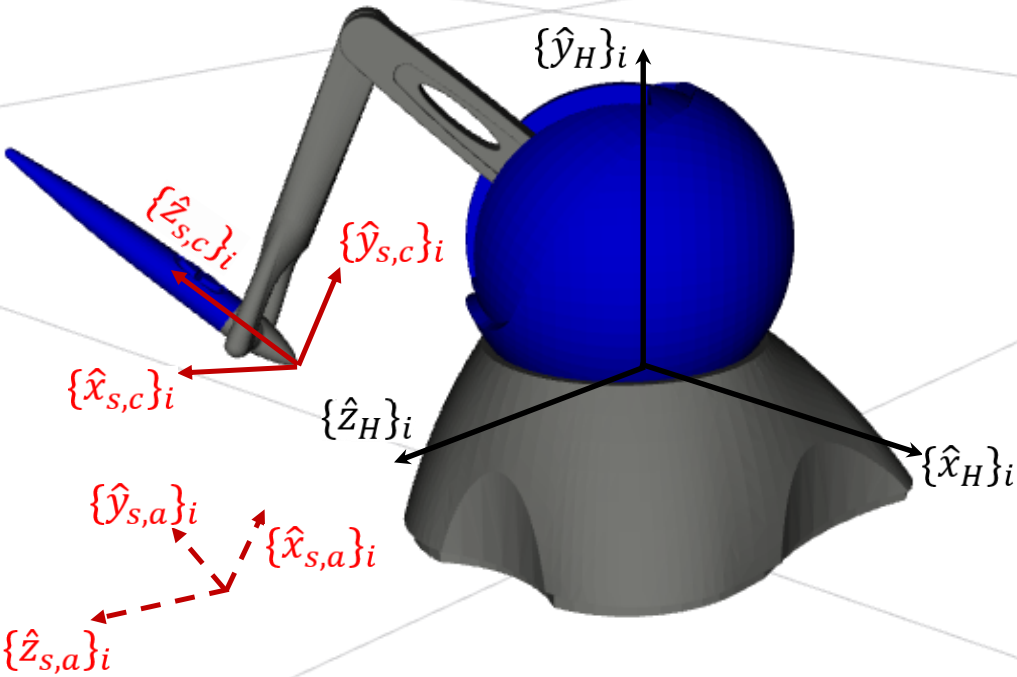}
        \caption{Coordinate frames of leader haptic device, with anchor frame indicated}
        \label{fig:haptic-frames}
    \end{subfigure}
    \begin{subfigure}{\linewidth}
        \centering
        \includegraphics[width=0.7\linewidth]{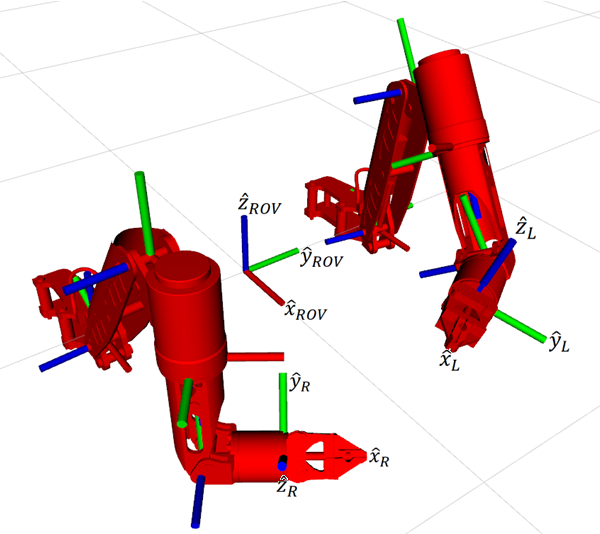}
        \caption{Coordinate frames of dual follower manipulators, visualized using RViz}
        \label{fig:manipulator-frames}
    \end{subfigure}
    \caption{Coordinate frames of haptic devices and manipulators, including anchor frames}
\end{figure}

\subsection{Commanding Manipulator Trajectory}

Teleoperation of the manipulators is commanded when only the proximal button, referred to as the manipulator button, on the corresponding stylus is engaged. In this proposed system, both follower manipulators can be controlled simultaneously or individually. When teleoperation of a follower serial manipulator is initiated, the haptic stylus pose at this instant, known as the anchor pose, is registered as \((^{H}\mathbf{T}_{s,a})_{i}\). The subsequent poses of the moving stylus, or current pose, while the button stays engaged is represented as \((^{H}\mathbf{T}_{s,c})_{i}\) as seen in Fig \ref{fig:haptic-frames}. The relationship between the anchor and current pose of the haptic stylus is as follows:
\begin{equation}
    (^{H}\mathbf{T}_{s,c})_{i} = ({}^{H}\mathbf{T}_{s,a})_{i} ({}^{s,a}\mathbf{T}_{s,c/a})_{i}\;,
\end{equation}
where \((^{s,a}\mathbf{T}_{s,c/a})_{i}\) represents the change in pose from the anchor frame to the current frame of the corresponding stylus, as represented in the anchor frame. This relative change in pose corresponds to the follower manipulator's relative change in pose using a similarity transformation by:
\begin{equation}
    (^{ee, a}\mathbf{T}_{s,c/a})_{i} = ({}^{ee, a}\mathbf{T}_{s,a})_{i} ({}^{s,a}\mathbf{T}_{s,c/a})_{i} ({}^{ee, a}\mathbf{T}_{s,a})^{-1}_{i}\;,
\end{equation}

\noindent where \(({ee,a})\) is the end-effector frame corresponding to the manipulator's pose when the manipulator button is initially pressed. \(({}^{ee, a}\mathbf{T}_{s,a})_{i}\) is computed using the following expression:
\begin{equation}
    (^{ee, a}\mathbf{T}_{s,a})_{i} = ({}^{ROV}\mathbf{T}_{ee, a})^{-1}_{i} ({}^{ROV}\mathbf{T}_{H})_{i} ({}^{H}\mathbf{T}_{s,a})_{i}\;,
\end{equation}

\noindent where \((^{ROV}\mathbf{T}_{ee,a})_{i}\) is the manipulator anchor pose as written in the vehicle base frame.

This transformed relative pose is essential to compute the pose of the teleoperated manipulator written in the base frame of the vehicle using the following expression:
\begin{equation}
    (^{ROV}\mathbf{T}_{ee,d})_{i} = ({}^{ROV}\mathbf{T}_{ee, a})_{i} ({}^{ee, a}\mathbf{T}_{s,c/a})_{i}\;, \label{eq:teleop-desired}
\end{equation}

\noindent where \((^{ROV}\mathbf{T}_{ee,d})_{i}\) is the end-effector desired pose $\mb{x}_{d,i}$ in the form of a \(4\times4\) homogeneous transformation. The desired pose is used in the resolved motion rate controller introduced in Section \ref{section:uvms-kinematics}.

\subsection{Vehicle Teleoperation}

Teleoperation of the vehicle occurs when both leader haptic devices have only the distal button on their stylus pressed. When the buttons are pressed, the initial positions of the styluses are registered and corresponding positions are rewritten as vectors denoting the relative position. A vector in the haptic base frame is represented component-wise in the axes of the same frame. Note that the rotation of the stylus pose is ignored for this mode. To command motion of the vehicle, the vectors of both devices must have the same dominant component, and both vectors must have a norm above a nominal distance. The condition for the vector norm is to create a dead-zone, which will prevent the vehicle from moving in unwanted directions when initiating vehicular teleoperation. While both leader device vectors share the same dominant component, they must also share the same direction to command a constant velocity along that direction. In the $\{ROV\}$ base frame, this translates to a linear velocity in the direction of its components. Additionally, the yaw of the vehicle can also be controlled when the two haptic vectors have the dominant \(\hat{\mb{z}}_{H,i}\) axes but are in opposite directions. In this case, the angular velocity of the vehicle in the \(\hat{\mb{z}}_{ROV}\) axes is commanded. 

\subsection{Control Architecture}

Figure \ref{fig:rr-schematic} displays the control architecture for bimanual teleoperation of a UVMS. 
The two haptic devices communicate their poses and button states in real-time to the teleoperation controller. When controlling the manipulators, the teleoperation controller communicates the desired poses to the resolved motion rate controller, which computes the joint velocities to determine and command joint positions of the follower manipulators to the joint PID controller in Gazebo. An RViz~\cite{rviz} visualization of the manipulator poses is also deployed simultaneously.
When controlling the vehicle, the teleoperation controller sends velocity commands directly to the vehicle PID controller in Gazebo. 

\section{Simulation Model}
\label{section:simulation-model}

To test the proposed teleoperational framework, the bimanual UVMS is imported into the Gazebo simulator, a 3D dynamic environment for modeling robots, actuators, and sensors in a realistic world \cite{koenig2004}. The physics engine is capable of modeling gravity, inertia, friction, and collision detection between objects. Self-collision detection is possible with Gazebo, but is not utilized in this paper. The actuation of the serial manipulators is achieved via joints which define the kinematic and dynamic relationship between links. Another important attribute of this simulator is that it is open-source and compatible with many custom plugins, including integration with ROS to control the robot. 

To simulate underwater vehicles in Gazebo, the UUV Simulator will be used, which contains custom Gazebo plugins for underwater objects such as thrusters and sensors \cite{manhaes2016}. In addition, the package is compatible with ROS nodes to allow for features such as dynamic position (DP) control of the vehicle, trajectory generation, and manipulator control. The package also expands the dynamic model to include hydrodynamics and hydrostatics such as buoyancy and drag forces.

The purpose of the simulation is to provide preliminary validation of the framework's performance in two different tasks. The first task is manual piloting of the vehicle, manipulators, and end-effectors to a desired pre-grasp pose. In this task, the ability to execute commands of all components of the robot is demonstrated. In addition, the teleoperation command, resolved motion rate controller command, and actual position of the end-effectors are compared to assess the accuracy of the framework. The second task involves grasping a small object using both manipulators simultaneously. This demonstrates precise, coordinated motion of the whole robot with just a single novice operator, as well as the ability of the simulator to model the physics of a multi-point grasp.

\subsection{Vehicle Model}
The RexROV is the platform utilized in this simulation. This vehicle was used because it is one of the default vehicles included in the UUV Simulator packages. In addition, the size and thruster configuration of the vehicle is similar to that of light-duty or work class ROVs such as the Saab Lynx or Cougar models \cite{saab}. However, the proposed framework can easily be generalized to any light-duty underwater vehicle, such as the BlueROV2. This is to be consistent with the goal of lowering the barrier to entry into UVMS control through compatibility with a variety of UVMS designs.

To control the vehicle pose, a combined PID position and velocity controller is implemented. This hybrid controller maintains an upright orientation of the vehicle by using position control to set both the roll and pitch of the vehicle to zero. The user cannot control these vehicle DoFs. The controller's velocity PID control allows the user to control the linear velocities and yaw rate of the vehicle. The controller combines the orientation control with the user's desired velocity control to publish a desired vehicle wrench. A thruster controller generated by the UUV Simulator package subscribes to the vehicle wrench and calculates the actuation (thruster) forces required to achieve this wrench. The thruster controller is automatically generated based on the geometry of the thruster placement and the thruster gains. 

The PID controller allows the user to send velocity commands to the vehicle in one of two ways. The default method is via gamepad controller or joystick, which is a common method of controlling small ROVs such as the BlueROV2. However, we also developed a novel vehicle control method that uses the haptic devices to generate desired vehicle commands when the user presses the vehicle button on both manipulators. This allows for a single operator to effortlessly switch between vehicle and manipulator control.

\subsection{Manipulator Model}

The framework provided can be used for any desired configuration or manipulator, with the only required modification being that of the kinematic model parameters to match the geometry of the manipulators. For example, the Reach Alpha by Reach Robotics \cite{reachrobotics} could be used instead of the simple design shown on the BlueROV2 in Fig.~\ref{fig:UVMS-frames}. 

The manipulators are controlled via a PID position controller for each joint, including the gripper joints. The desired end-effector poses from the teleoperation tracking are calculated independently based on the transformation between the user's current pose and the anchor pose set when the manipulator button is engaged. Since there are only two buttons on each haptic device, the gripper is actuated when the user presses both buttons at the same time. The desired poses are sent to a resolved motion rate controller which calculates the joint commands to achieve the desired pose. As such, there are two types of error to be considered: pose error of the end-effector and joint position errors of the PID controller for each joint.



\section{Simulation Results}
\label{section:simulation-results}

\subsection*{Task 1: Motion Control of UVMS}

The first simulation task is to implement the proposed framework for real-time teleoperational control of the entire UVMS. In this task, the robot will move into a pre-grasp position to interact with a subsea panel, which is a common task required for intervention ROVs. This task has two modes: the vehicle motion and the manipulation mode. In the first mode, the vehicle is piloted into position using both haptic devices, and the manipulators cannot be controlled. In the second mode, the manipulators and grippers are controlled independently, and the vehicle cannot be controlled. The operator can freely switch between either mode as needed to move the end-effectors into the desired pre-grasp pose.

\begin{figure*}
    \captionsetup[subfigure]{justification=centering}
    \centering
    \begin{subfigure}{0.48\linewidth}
        \centering
        \includegraphics[width=\linewidth]{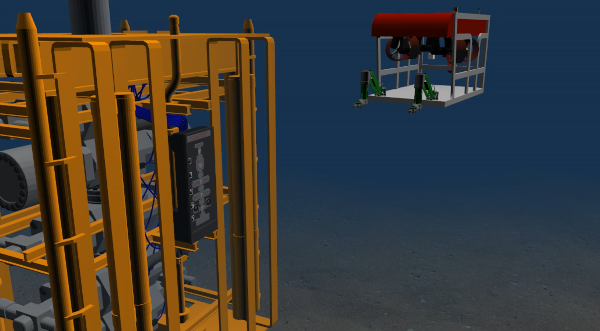}
        \caption{3\textsuperscript{rd} person view of Task 1 simulation. The UVMS is spawned and vehicle and manipulator controllers are enabled}
        \label{fig:task1-spawn}
    \end{subfigure}
    \begin{subfigure}{0.48\linewidth}
        \centering
        \includegraphics[width=\linewidth]{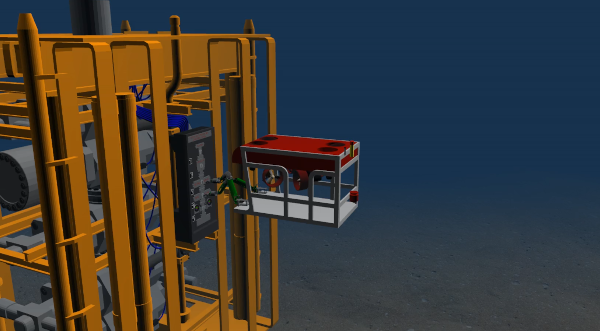}
        \caption{3\textsuperscript{rd} person view of Task 1 simulation. The UVMS is piloted into the pre-grasp position}
        \label{fig:task1-pilot}
    \end{subfigure}
    \begin{subfigure}{0.48\linewidth}
        \centering
        \includegraphics[width=\linewidth]{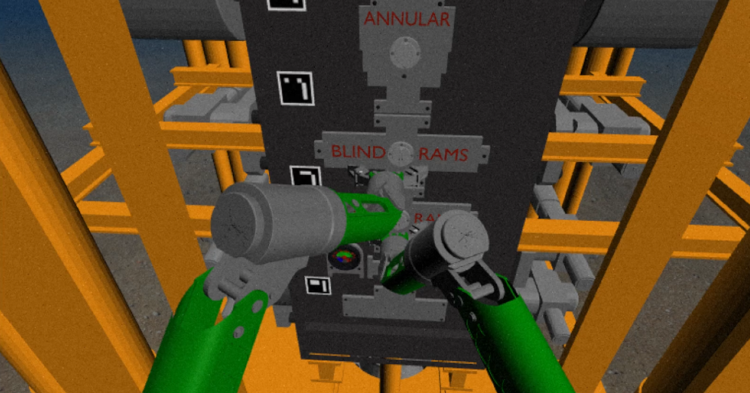}
        \caption{Robot camera view of Task 1 simulation. The manipulators are moved into final position to interact with the subsea panel}
        \label{fig:task1-manipulators}
    \end{subfigure}
    \begin{subfigure}{0.48\linewidth}
        \centering
        \includegraphics[width=\linewidth]{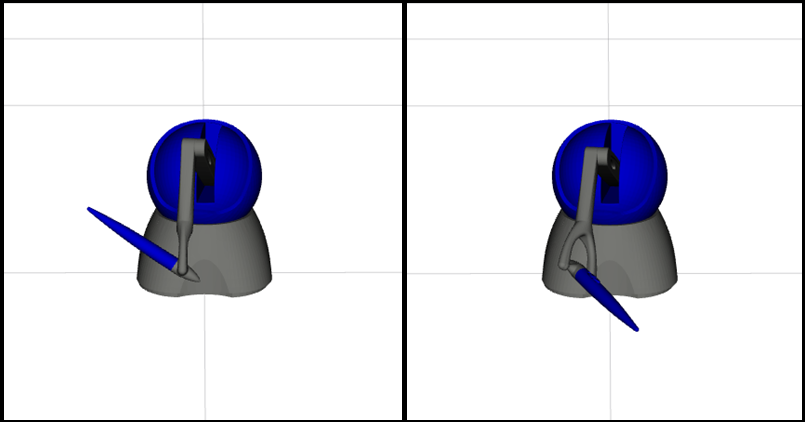}
        \caption{RViz visualization of user input on the haptic devices during Task 1 simulation}
        \label{fig:task1-haptic}
    \end{subfigure}
    \caption{Task 1 demonstration: controlling vehicle and manipulators using haptic devices}
    \label{fig:task1}
\end{figure*}

Figure~\ref{fig:task1} summarizes the framework implementation over the duration of the simulation.  First, the UVMS is spawned and the thruster and manipulator controllers are activated. Once this occurs, both the vehicle and the manipulators will remain stationary until the user gives a command via the haptic devices, as shown in Fig. \ref{fig:task1-spawn}. Next, the vehicle velocity control is demonstrated using the haptic devices to pilot the robot into an ideal position. This is done by pressing the vehicle button on each haptic device and using the haptic device pens to move the vehicle in strafing motions or rotational (yaw) motion. Figure \ref{fig:task1-pilot} shows the vehicle after it has been piloted to its pre-grasp position. Once the vehicle is in the desired position, the manipulators are engaged and moved to the desired position by pressing the manipulator button on the pen. The grippers are opened and closed by pressing both buttons on the corresponding pen. During this task, the vehicle camera view is used to provide a realistic simulated view of what the operator would see; the default third person view can be used for assistance in positioning, but it is important to note that in a real teleoperational task this would likely not be available to the operator. The camera view used in the simulation is shown in Fig. \ref{fig:task1-manipulators}. Finally, Fig. \ref{fig:task1-haptic} shows the haptic device motion during the simulation, indicating the actual commands given by the operator.

\begin{figure*}
    \captionsetup[subfigure]{justification=centering}
    \centering
    \begin{subfigure}{0.48\linewidth}
        \centering
        \includegraphics[width=\linewidth]{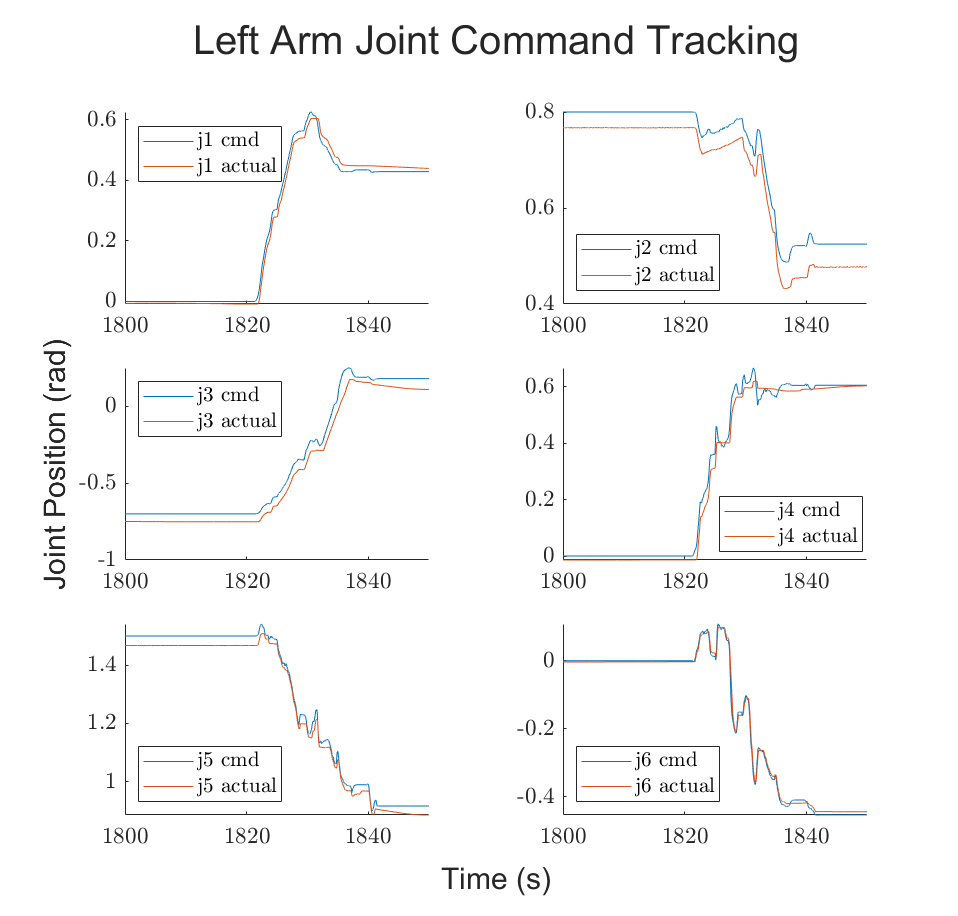}
        \caption{Comparison of individual left arm joint commanded and actual position during Task 1 simulation}
        \label{fig:task1-jointL}
    \end{subfigure}
    \begin{subfigure}{0.48\linewidth}
        \centering
        \includegraphics[width=\linewidth]{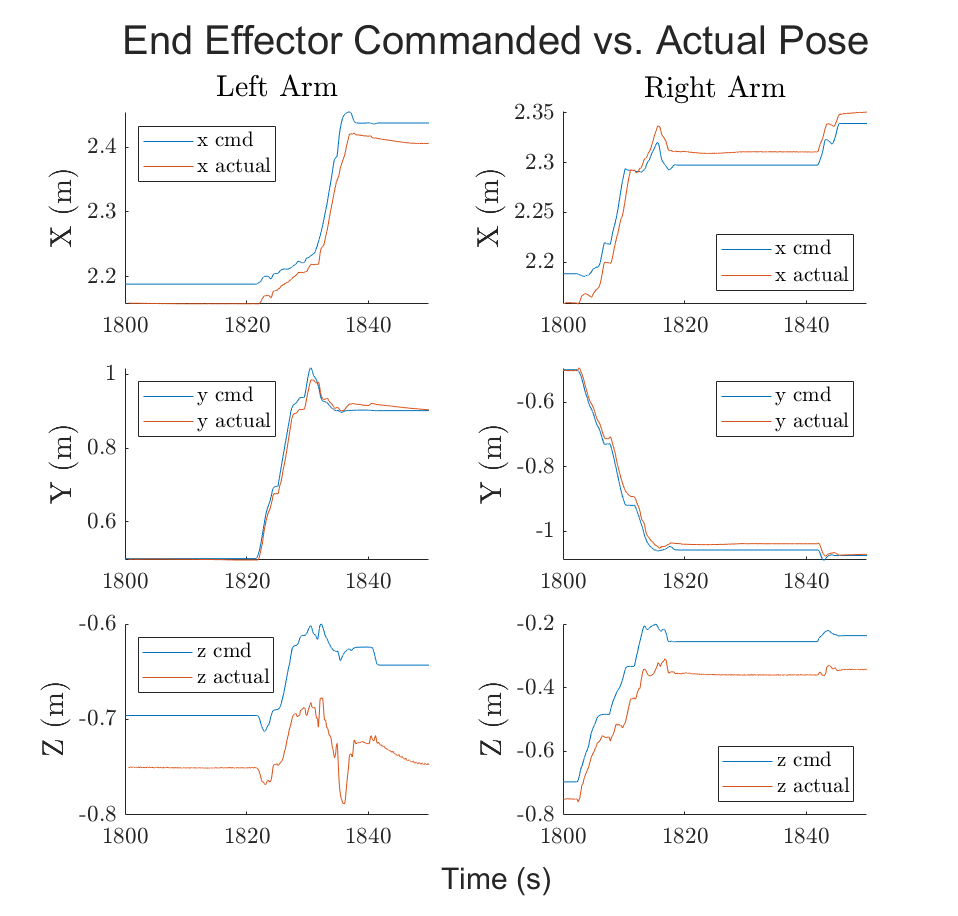}
        \caption{Comparison of end-effector commanded and actual position during Task 1 simulation}
        \label{fig:task1-trajectory}
    \end{subfigure}
   \caption{Task 1 manipulator performance analysis}
    \label{fig:task1-analysis}
\end{figure*}

The error associated with Task 1 is shown in Fig. \ref{fig:task1-analysis}. Figure \ref{fig:task1-jointL} shows the joint commanded and actual positions of each joint for the left manipulator. The PID controller error is very small for most joints but more pronounced in joint 2 and, to a lesser extent, joint 3. This makes sense because the second joint is the one which bears the greatest load in the default configuration. This error is not significant because the operator can easily compensate using visual feedback to get the manipulators into the desired position. The PID gains of these joints can also be adjusted to reduce this error if desired.

The commanded and actual end-effector positions in the ROV base frame are shown in Fig. \ref{fig:task1-trajectory}. The error is greatest in the Z direction, small in the X direction, and very small in the Y direction. This is consistent with the joint 2 errors, representing the robot sagging under load and causing primarily the Z position to be lower in actuality than commanded. As before, this can be compensated for manually by the operator or by adjusting the PID gains.

\subsection*{Task 2: Bimanual Manipulation of an Object}

The second task uses the framework to perform a more challenging task to grasp an object using both manipulators simultaneously. This task is similar in concept to the previous task and uses the same two modes, but with a greater emphasis on precise manipulation and the physics of grasping. The object in this case is a simple rectangular box spawned into Gazebo for demonstration purposes. In this task, the object is grasped with both manipulators, which requires coordination and precision. The fact that this can be grabbed by a novice operator demonstrates the usefulness of the proposed framework. Another advantage to using bimanual grasping is that the object can be stabilized using both manipulators. A single gripper would have difficulty grabbing such an object because of its smooth surface, allowing the object to slip both length-wise as well as laterally out of the gripper. However, using both manipulators at once allows the object to be caged by both grippers, forming a secure grasp to prevent lateral motion. 

\begin{figure}
    \centering
    \includegraphics[width=\linewidth]{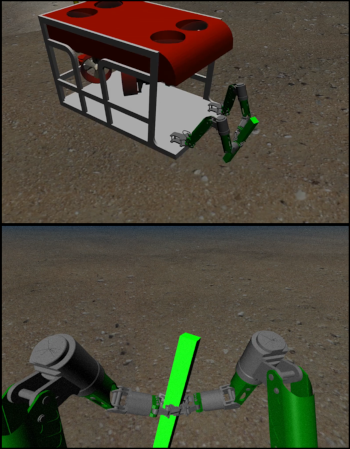}
    \caption{Task 2 demonstration: both manipulators are engaged to grasp a rectangular object, shown in green}
    \label{fig:task2}
\end{figure}

Figure \ref{fig:task2} shows Task 2 in execution. At top, the third-person view of the robot shows the object being lifted off the seafloor. At bottom, the robot view is shown, which is what the operator will typically use in a real teleoperational task. Once grasped, the object can be transported by moving the vehicle itself or by moving the manipulators; however, the operator must be careful to move the object with both manipulators in coordination, or the grasp will fail.

\section{Conclusion}
\label{section:conclusion}

In this paper, an open-source bimanual teleoperation framework for a light duty UVMS is introduced. The novelty of the framework lies in the use of low-cost and readily available haptic devices to control both the vehicle and manipulator motions. The requisite UVMS kinematics are defined and a resolved motion rate controller is given to independently control both manipulators. The teleoperational mapping between the haptic devices and the desired manipulator pose is also derived. The framework is validated in simulation in two tasks. The first task is to demonstrate the motion control of the vehicle, manipulators, and grippers using the haptic devices only. The performance of the robot is assessed by analyzing the joint and end-effector position errors. The second task is to grasp an object using both manipulators simultaneously, which demonstrates the precise and coordinated control of the UVMS using the proposed framework.

Future development of the framework includes coordination between macro-level and micro-level control of the UVMS. The manipulator and vehicle control presented in the current framework constitutes micro-level control. Macro-level control would encompass simultaneous localization and mapping (SLAM) and autonomous execution of user-generated waypoint trajectories. Such a framework would be very valuable for completing more complex tasks and for lowering the barrier to entry into underwater robotics. In addition, the incorporation of haptic feedback to assist the operator could provide useful capabilities such as virtual fixtures to guide the operator through a pre-determined task or indicate contact with an object. Finally, testing the framework on an experimental platform would demonstrate the real-world utility of this framework.

\bibliographystyle{IEEEtran}
\bibliography{formatting/sources-teleop, formatting/sources-uvms}

\end{document}